\documentclass[]{llncs}
\usepackage[dvipdfmx]{graphicx}
\usepackage{amsmath,amssymb}
\usepackage{color}
\usepackage[width=122mm,left=12mm,paperwidth=146mm,height=193mm,top=12mm,paperheight=217mm]{geometry}

\usepackage{array}
\usepackage{booktabs}
\usepackage{comment}
\usepackage{multirow}
\usepackage{paralist}
\usepackage{url}
\graphicspath{{fig/}}

\begin{document}
\pagestyle{headings}
\mainmatter

\title{Parallel Grid Pooling for Data Augmentation} 

\author{{Akito Takeki, Daiki Ikami, Go Irie$^\dagger$, and Kiyoharu Aizawa}}
\institute{\url{{takeki, iakmi, aizawa}@hal.t.u-tokyo.ac.jp}, The University of Tokyo\\
	$^\dagger$ \url{goirie@ieee.org}, NTT Communications Science Laboratories
}

\maketitle

\begin{abstract}
Convolutional neural network (CNN) architectures utilize downsampling layers, which restrict the subsequent layers to learn spatially invariant features while reducing computational costs.
However, such a downsampling operation makes it impossible to use the full spectrum of input features.
Motivated by this observation, we propose a novel layer called parallel grid pooling (PGP) which is applicable to various CNN models.
PGP performs downsampling without discarding any intermediate feature.
It works as data augmentation and is complementary to commonly used data augmentation techniques.
Furthermore, we demonstrate that a dilated convolution can naturally be represented using PGP operations, which suggests that the dilated convolution can also be regarded as a type of data augmentation technique.
Experimental results based on popular image classification benchmarks demonstrate the effectiveness of the proposed method.
Code is available at: \url{https://github.com/akitotakeki/pgp-chainer}.
\keywords{Data Augmentation, Dilated Convolution, Convolutional Neural Networks, Image Classification}
\end{abstract}

\section{Introduction}
\label{section:intro}
Deep learning using convolutional neural networks (CNNs) has achieved excellent performance for a wide range of tasks, such as image recognition~\cite{he2016deep,huang2016densely,hu2017squeeze}, object detection~\cite{liu2016ssd,redmon2017yolo9000,lin2017focal}, and semantic segmentation~\cite{chen2016deeplab,zhao2017pyramid}.
Most CNN architectures utilize spatial operation layers for downsampling.
Spatial downsampling restricts subsequent layers in a CNN to learn spatially invariant features and reduces computational costs.
Modern architectures often use multi-stride convolutional layers, such as a $3\times3$ or $1\times1$ convolution with a stride of 2~\cite{he2016deep,hu2017squeeze,gross2016training,zagoruyko2016wide,xie2017aggregated,he2016identity} or $2\times2$ average pooling with a stride of 2~\cite{huang2016densely}.
The other methods for downsampling have been proposed~\cite{zeiler2013stochastic,graham2014fractional,lee2016generalizing,zhai2016s3pool}.

\begin{figure}[tb]
	\centering
	\begin{tabular}{cc}
	\includegraphics[height=.15\textheight]{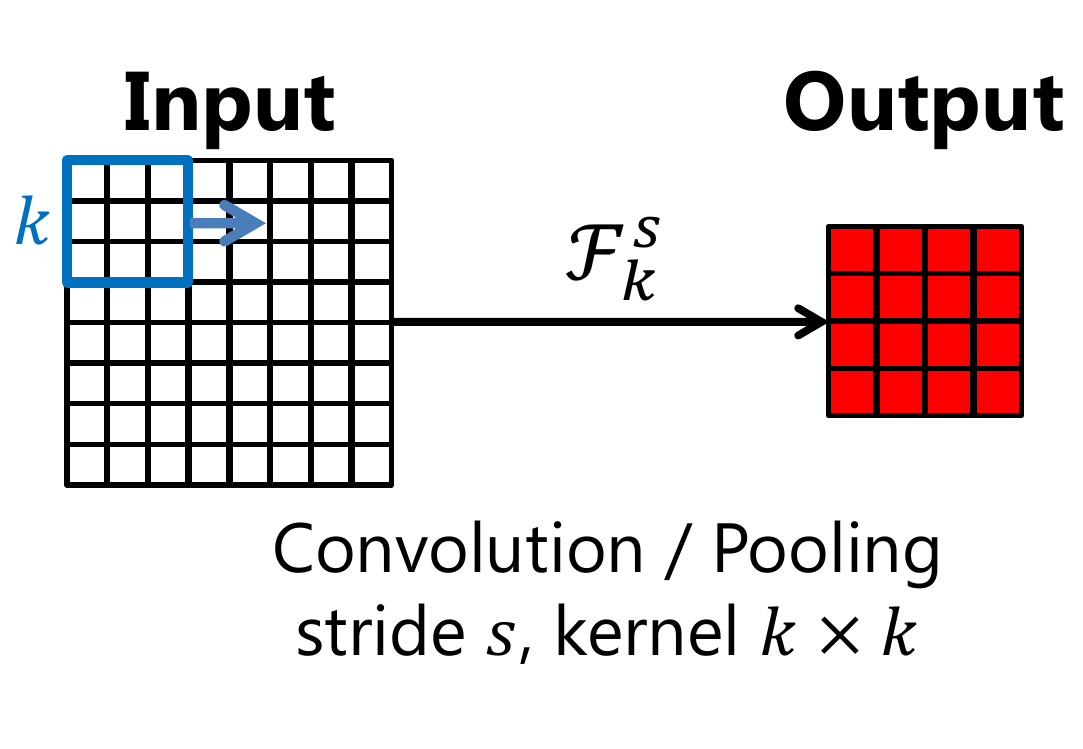} & \includegraphics[height=.15\textheight]{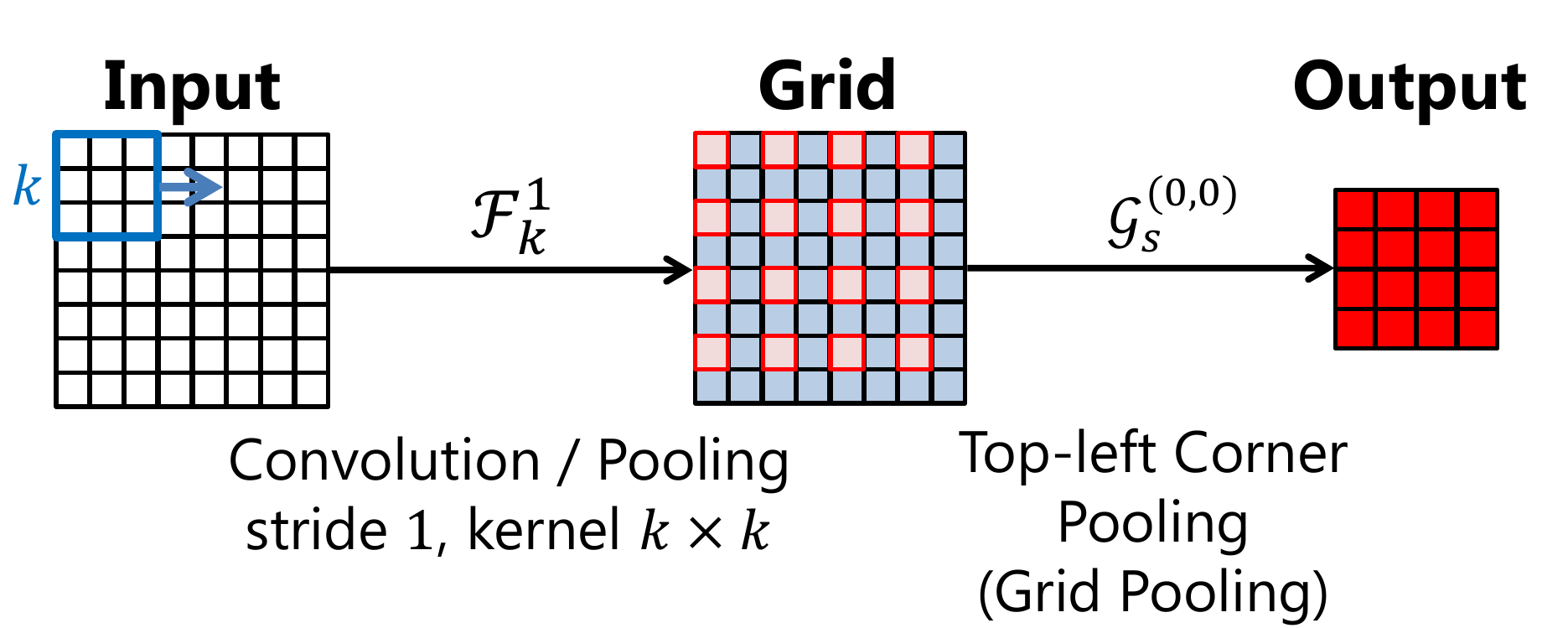}\\
	(a) & (b) \\
	\end{tabular}
	\caption{A two step view of a spatial operation (e.g. convolution or pooling) $\mathcal{F}^s_k$ with a kernel size of $k\times k$ and stride of $s\times s~(s\geq2)$. (a) is equivalent to (b).}
	\label{fig:2step}
\end{figure}

One drawback of these downsampling is that it makes impossible to use the full spectrum of input features, and most previous works overlook the importance of this drawback. 
To clarify this issue, we provide an example of such a downsampling operation in Fig.~\ref{fig:2step}.
Let $\mathcal{F}_k^s$ be a spatial operation (e.g., convolution or pooling) with a kernel size of $k\times k$ and stride of $s\times s~(s\geq2)$.
In the first step, $\mathcal{F}^1_k$ is performed on the input and yields an intermediate output.
In the second step, the intermediate output is split into an $s\times s$ grid pattern and downsampled by selecting the coordinate $(0, 0)$ (top-left corner) from each grid, resulting in an output feature that is $s$ times smaller than the input feature. 
In this paper, the second step downsampling operation $\mathcal{G}^{(0,0)}_s$ is referred to as \textbf{grid pooling (GP)} for the coordinate $(0, 0)$.
From this two-step example, one can see that the typical downsampling operation $\mathcal{F}^s_k$ utilizes only $1/s^2$ of the intermediate output and discards the remaining $1-1/s^2$.

Motivated by the observation, we propose a novel layer for CNNs called the \textbf{parallel grid pooling (PGP)} layer (Fig.~\ref{fig:pgp}).
PGP is utilized to perform downsampling without discarding any intermediate feature and can be regarded as a data augmentation technique.
Specifically, PGP splits the intermediate output into an $s\times s$ grid pattern and performs a grid pooling for each of the coordinates $(i, j)$~$(0\leq i\leq s-1, 0\leq j\leq s-1)$ in the grid.
This means that PGP transforms an input feature into $s^2$ feature maps, downsampled by $1/s^2$.
The layers following PGP compute $s^2$ feature maps shifted by several pixels in parallel, which can be interpreted as 
data augmentation in the feature space.

Furthermore, we demonstrate that dilated convolution~\cite{chen2016deeplab,yu2017dilated,oord2016wavenet,kalchbrenner2016neural} can naturally be decomposed into a convolution and PGP.
In general, a dilated convolution is considered as an efficient operation for spatial (and temporal) information, and achieves the state-of-the-art performance for various tasks, such as semantic segmentation~\cite{chen2016deeplab,yu2015multi,chen2018encoder}, speech recognition~\cite{oord2016wavenet}, and natural language processing~\cite{kalchbrenner2016neural,yang2017improved}.
In this sense, we provide a novel interpretation of the dilated convolution operation as a data augmentation technique in the feature space.

The proposed PGP layer has several remarkable properties.
PGP can be implemented easily by inserting it after a multi-stride convolutional and/or pooling layer without altering the original learning strategy.
PGP does not have any trainable parameters and can even be utilized for test-time augmentation (i.e., even if a CNN was not trained using PGP layers, the performance can still be improved by inserting PGP layers into the pretrained CNN at testing time).

We evaluate our method on three standard image classification benchmarks: CIFAR-10, CIFAR-100~\cite{hinton2007learning}, SVHN~\cite{netzer2011reading}, and ImageNet~\cite{russakovsky2015imagenet}.
Experimental results demonstrate that PGP can improve the performance of recent state-of-the-art network architectures and is complementary to widely used data augmentation techniques (e.g., random flipping, random cropping, and random erasing).

The major contributions of this paper are as follows:
\begin{itemize}
\item We propose PGP, which performs downsampling without discarding any intermediate feature information and can be regarded as a data augmentation technique. PGP is easily applicable to various CNN models without altering their learning strategies and the number of parameters.
\item We demonstrate that PGP is implicitly used in a dilated convolution operation, which suggests that dilated convolution can also be regarded as a type of data augmentation technique.
\end{itemize}

\section{Related Works}
\label{section:related}
PGP is most closely related to data augmentation.
In the context of training a CNN, data augmentation is a standard regularization technique that is used to avoid overfitting and artificially enlarge the size of a training dataset.
For example, applying crops, flips, and various affine transforms in a fixed order or at random is a widely used technique in many computer vision tasks.
Data augmentation techniques can be divided into two categories: data augmentation in the image space and data augmentation in the feature space.

\textbf{Data augmentation in the image space:} 
AlexNet~\cite{krizhevsky2012imagenet}, which achieves state-of-the-art performance on the ILSVRC2012 dataset, applies random flipping, random cropping, color perturbation, and adding noise based on principal component analysis.
TANDA~\cite{ratner2017learning} learns a generative sequence model that produces data augmentation strategies using adversarial techniques on unlabeled data.
The method in \cite{hauberg2016dreaming} proposes to train a class-conditional model of diffeomorphisms by interpolating between nearest-neighbor data.
Recently, training techniques utilizing randomly masked images of rectangular regions, such as random erasing~\cite{zhong2017random} and cutout~\cite{devries2017improved}, have been proposed. 
Using these augmentation techniques, recent methods have achieved state-of-the-art results for image classification~\cite{yamada2018shakedrop,real2018regularized}.
Mix-up~\cite{zhang2017mixup} and BC-learning~\cite{yuji2017between} generate between-class images by mixing two images belonging to different classes with a random ratio.

\textbf{Data augmentation in the feature space:} 
Some techniques generate augmented training sets by interpolating~\cite{chawla2002smote} or extrapolating~\cite{devries2017dataset} features from the same class.
DAGAN~\cite{fawzi2016adaptive} utilizes conditional generative adversarial networks to conduct feature space data augmentation for one-shot learning.
A-Fast-RCNN~\cite{wang2017fast} generates hard examples with occlusions and deformations using an adversarial learning strategy in feature space.
Smart augmentation~\cite{lemley2017smart} trains a neural network to combine training samples in order to generate augmented data.
As another type of feature space data augmentation, AGA~\cite{dixit2017aga} and Fatten~\cite{liu2018feature} perform augmentation in the feature space by using properties of training images (e.g., depth and pose).

While our PGP performs downsampling without discarding any intermediate feature, it splits the input feature map into branches, which will be discussed in detail in Sec.~\ref{subsection:pgp}.
The operations following PGP (e.g., convolution layers or fully connected layers) compute slightly shifted $s^2$ feature maps, which can be interpreted as a novel data augmentation technique in the feature space.
In addition, dilated convolution can be regarded as data augmentation technique in the feature space, which will be discussed in detail in Sec.~\ref{subsection:dil_ide}.

\section{Proposed Method}
\label{section:method}

\subsection{Parallel Grid Pooling (PGP)}
\label{subsection:pgp}
\begin{figure}[tb]
	\centering
	\includegraphics[width=.9\textwidth]{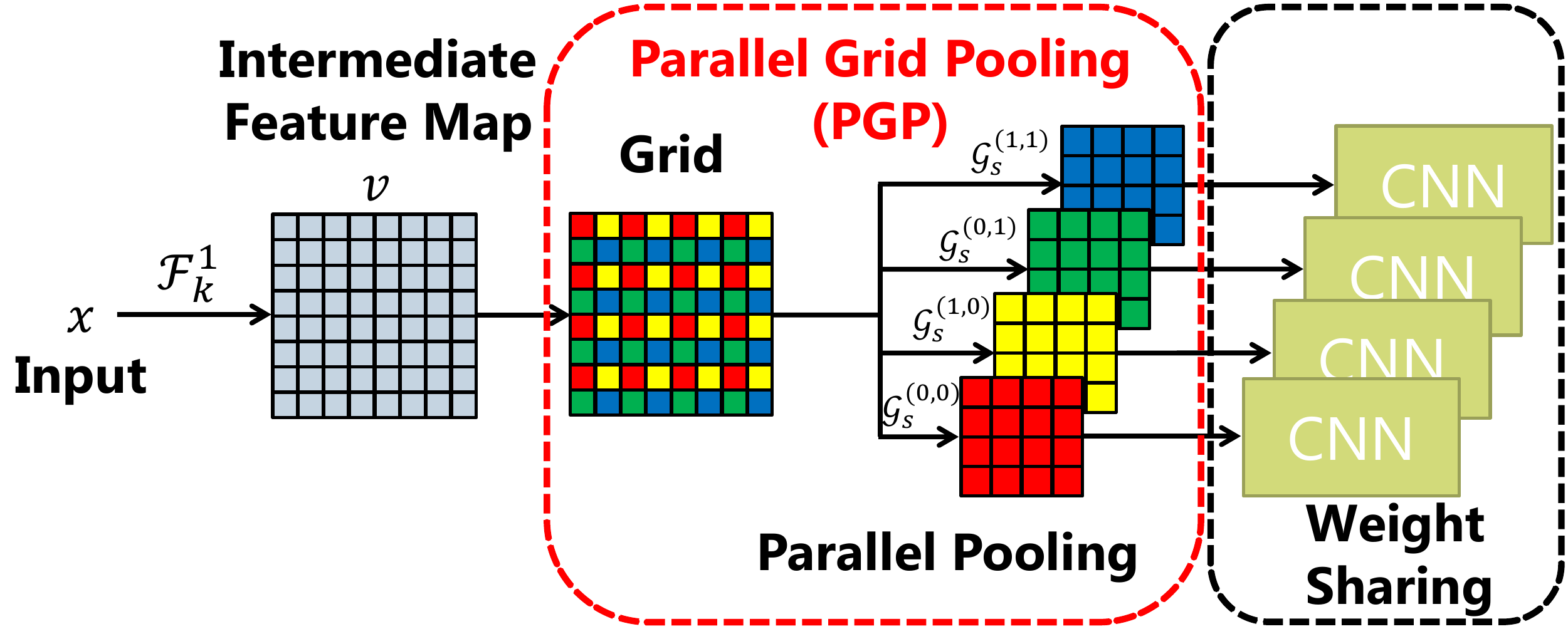}
	\caption{An overview of Parallel Grid Pooling (PGP)}
	\label{fig:pgp}
\end{figure}

An overview of PGP is illustrated in Fig.~\ref{fig:pgp}.
Let $x\in \mathbb{R}^{b\times c\times h\times w}$ be an input feature map, where $b$, $c$, $h$, and $w$ are the number of batches, number of channels, height and width, respectively.
 $\mathcal{F}^s_k$ is a spatial operation (e.g. a convolution or a pooling) with a kernel size of $k\times k$ and stride of $s\times s~(s\geq2)$.
The height and width of the output feature map are $h/s$ and $w/s$, respectively.

$\mathcal{F}^s_k$ can be divided into two steps (Fig.~\ref{fig:2step}).
In the first step, $v=\mathcal{F}^1_k\left( x\right)$ is performed, which is an operation that makes full use of the input feature map information with a stride of one, producing an intermediate feature map $v$.
Then, $v$ is split into $w/s\times h/s$ blocks of size $s\times s$.
Let the coordinate in each grid be $(i, j)$~$(0\leq i\leq s-1, 0\leq j\leq s-1)$.
In the second step, $v$ is downsampled by selecting the coordinate $(0, 0)$ (top-left corner) in each grid.
Let this downsampling operation $\mathcal{G}^s_{\left( i,j\right)}$ of selecting the coordinate $(i, j)$ in each grid be called grid pooling (GP).
To summarize, given a kernel size of $k\times k$ and stride of $s\times s~(s\geq2)$, the multi-stride spatial operation layer is represented as
\begin{eqnarray}
\label{eqn:classic}
y&=&\mathcal{F}^s_k\left( x\right) \nonumber \\ 
&=&\mathcal{G}^s_{\left( 0,0\right)} \left( \mathcal{F}^1_k\left( x\right)\right) 
\end{eqnarray}

Our proposed PGP method is defined as:
\begin{eqnarray}
\label{eqn:pgp}
\centering
\left( y_{\left( 0,0\right) }, y_{\left( 0,1\right) }, \ldots , y_{\left( s-1,s-1\right) }\right) &=&\left( \mathcal{G}^s_{\left( 0,0\right)}\left( x\right), \mathcal{G}^s_{\left( 0,1\right)}\left( x\right), \ldots, \mathcal{G}^s_{\left( s-1,s-1\right)}\left( x\right) \right)\nonumber \\
&=&\mathrm{PGP}\left(x\right)
\end{eqnarray}
The conventional multi-stride operation uses $y_{\left( 0,0\right)}$ which consists of only $1/s^2$ of the intermediate output and discards the remaining $1-1/s^2$ of the feature map.
In contrast, PGP retains all possible choices of grid poolings, meaning PGP makes full use of the feature map without discarding the $1-1/s^2$ portion of the intermediate output in the conventional method.

A network architecture with a PGP layer can be viewed as one with $s^2$ internal branches by performing grid pooling in parallel.
Note that the weights (i.e., parameters) of the following network are shared between all branches, hence there is no additional parameter throughout the network.
PGP performs downsampling while maintaining the spatial structure of the intermediate features, thereby producing output features shifted by several pixels.
As described above, because the operations in each layer following PGP share the same weights across branches, the subsequent layers in each branch are trained over a set of slightly shifted feature maps simultaneously.
This works as data augmentation; with PGP, the layers are trained with $s^2$ times larger number of mini-batches compared to the case without PGP.

\subsection{PGP vs. Dilated Convolution.}
\label{subsection:dil_ide}
We here discuss the relationship between PGP and dilated convolution~\cite{chen2016deeplab,yu2015multi}.
Dilated convolution is a convolutional operation applied to an input with a defined gap. 
Dilated convolution is considered to be an operation that efficiently utilizes spatial and temporal information~\cite{chen2016deeplab,oord2016wavenet,kalchbrenner2016neural,yu2015multi,Yu2017}.
For brevity, let us consider one-dimensional signals.
Dilated convolution is defined as:
\begin{equation}
\label{eqn:dilation}
y\left[ i\right] =\sum_{l=1}^{L}x\left[ i+rl\right] u\left[ l\right],
\end{equation}
where $x[i]$ is an input feature, $y[i]$ is an output feature, $u[l]$ is a convolution filter of kernel size $L$, and $r$ is a dilation rate.
A dilated convolution is equivalent to a regular convolution when $r=1$. 
Let $i=rj+k~(0\leq k\leq r-1)$.
Then, the output $y\left[rj+k\right]$ is
\begin{eqnarray}
y\left[ rj+k\right] &=&\sum_{l=1}^{L}x\left[rj+k+rl\right] u\left[ l\right] \\
&=&\sum_{l=1}^{L}x\left[ r\left( j+l\right) +k\right] u\left[ l\right].
\end{eqnarray}
This suggests that the output $y\left[i\right]$, where $i\equiv k \pmod r$, can be obtained from the input $\lbrace x\left[i\right]~|~i\equiv k\pmod r\rbrace$.
To summarize, an $r$-dilated convolution is equivalent to an operation where the input is split into $r$ branches and convolutions are performed with the same kernel $u$, then spatially rearranged into a single output.

\begin{figure}[tb]
	\centering
	\includegraphics[width=.8\textwidth]{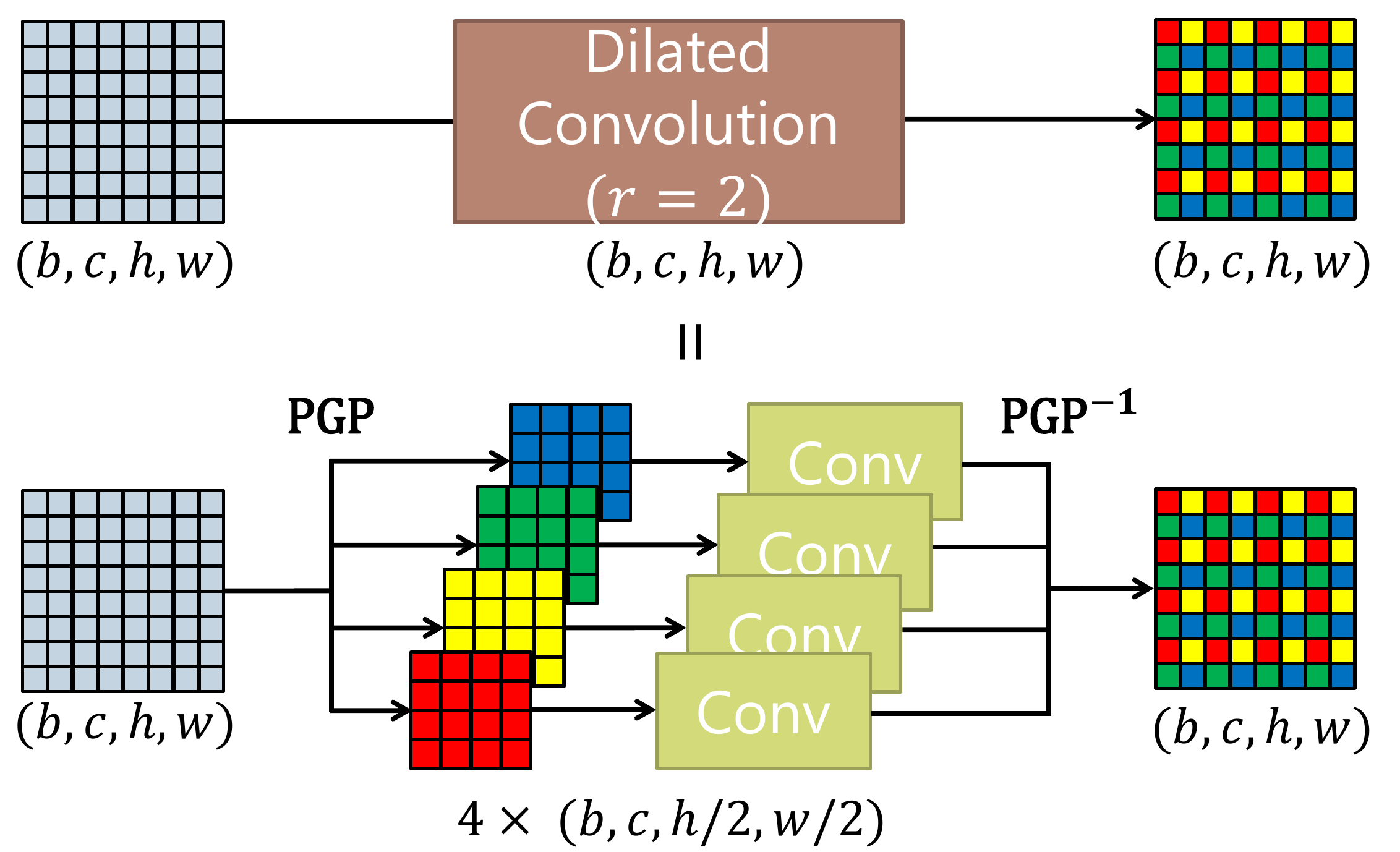}
	\caption{PGP is implicitly used by dilated convolution.}
	\label{fig:dilation}
\end{figure}

A dilated convolution in a two-dimensional space is illustrated in Fig.~\ref{fig:dilation} (in case $r=2$).
The $r$-dilated convolution downsamples an input feature with 
an interval of $r$ pixels, then produces $r^2$ branches as outputs.
Therefore, the dilated convolution is equivalent to a following three-step operation: $\mathrm{PGP}$, a convolution sharing the same weight, and inverse of PGP ($\mathrm{PGP}^{-1}$) which rearranges the cells to their original positions.
For example, let the size of the input features be $x=(b,c,h,w)$, where $b$, $c$, $h$, and $w$ are the number of batches, number of channels, height, and width, respectively. 
Using $\mathrm{PGP}$, the input feature $x$ is divided into $r^2\times(b, c, h/r, w/r)$, subsequently a convolutional filter (with shared weight parameters) is applied to each branch in parallel (Fig.~\ref{fig:dilation}).
Then, the intermediate $r^2$ feature maps are rearranged by $\mathrm{PGP}$, resulting the same size $(b,c,h,w)$ as the input by $\mathrm{PGP}^{-1}$.
In short, PGP is embedded in a dilated convolution, which suggests that the success of dilated convolution can be attributed to data augmentation in the feature space.

\subsection{Architectural Differences between dilated convolution and PGP}
\label{subsection:dil_dif}
\begin{figure}[tb]
	\centering
	\begin{tabular}{c}
		\includegraphics[width=.9\textwidth]{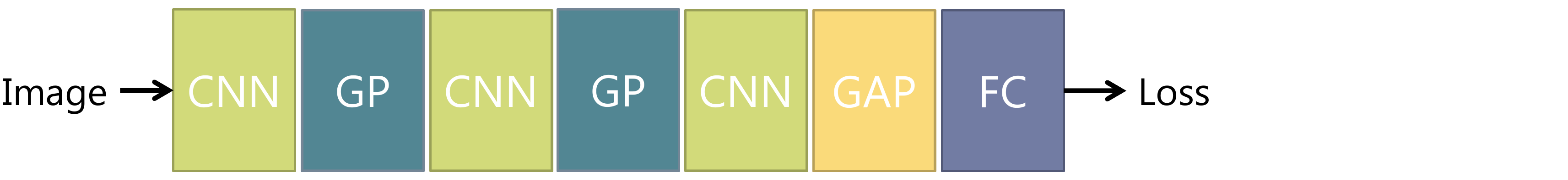} \\
		(a) Base CNN (Base-CNN) \\
		\includegraphics[width=.9\textwidth]{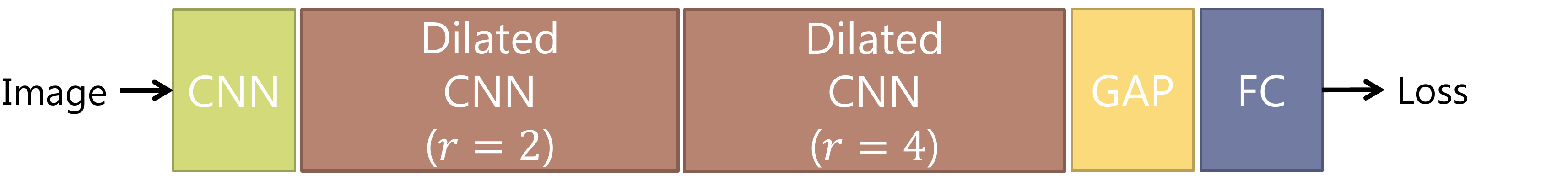} \\
		(b) CNN with dilated convolution (DConv-CNN)\\
		\includegraphics[width=.9\textwidth]{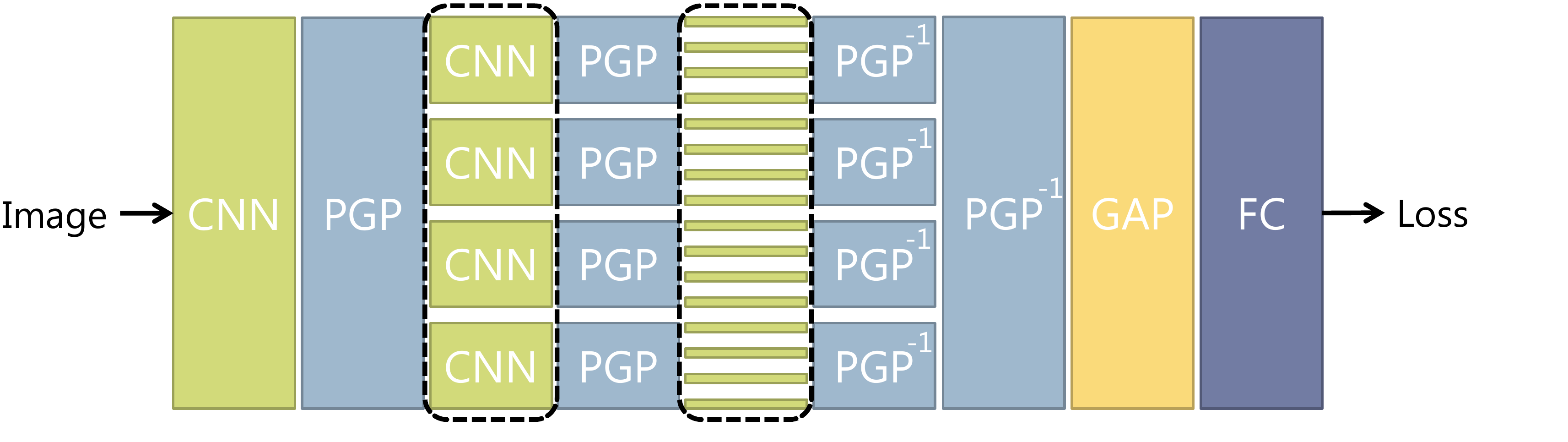} \\
		(c) DConv-CNN represented by PGP \\
		\includegraphics[width=.9\textwidth]{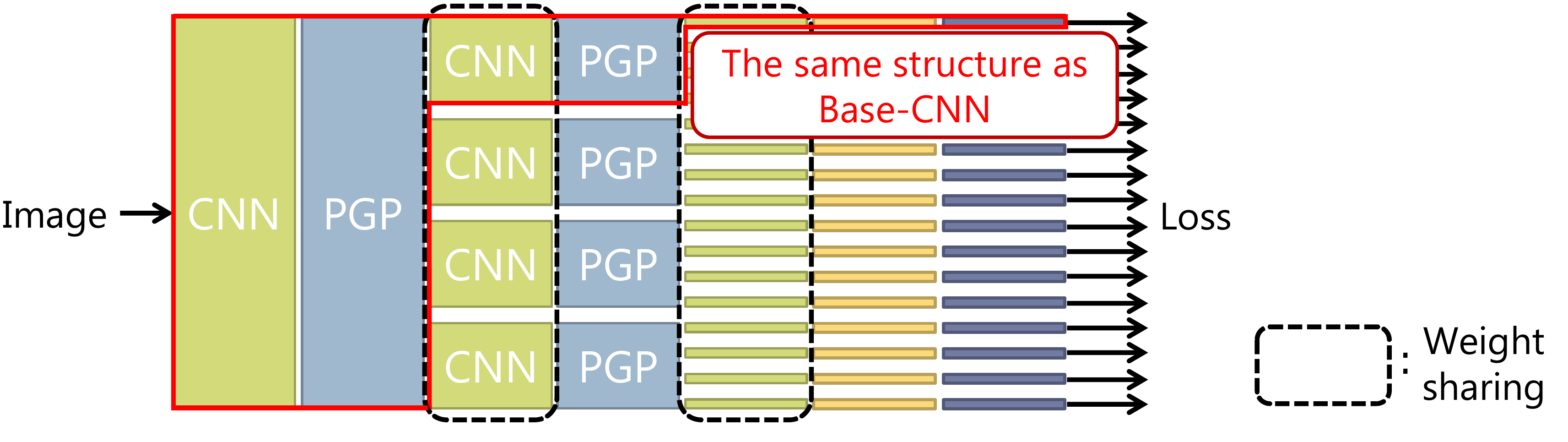} \\
		(d) CNN with PGP (PGP-CNN)
	\end{tabular}
	\caption{Comparison between DConv-CNN and PGP-CNN. As mentioned in Section~\ref{subsection:pgp}, Base-CNN is expressed as the architecture using GP~(a). (b) is equivalent to (c).}
	\label{fig:compare}
\end{figure}
The structure of the network model with PGP is better for learning than that with dilated convolutions.
For example, consider the cases where two downsampling layers in a base CNN (Base-CNN) are replaced by either dilated convolution (Fig.~\ref{fig:compare}(b)) or PGP (Fig.~\ref{fig:compare}(d)).
According to \cite{chen2016deeplab}, the conventional multi-stride downsampling layer and all subsequent convolution layers can be replaced with dilated convolution layers with two times dilation rate (Fig.~\ref{fig:compare}(b)).
As mentioned in Section~\ref{subsection:dil_ide}, a dilated convolution is equivalent to ($\mathrm{PGP}+ \mathrm{Convolution} +\mathrm{PGP}^{-1}$).
When $n$ $r$-dilated convolutions are stacked, they are expressed by ($\mathrm{PGP}+ \mathrm{Convolution} \times n +\mathrm{PGP}^{-1}$).
This means that each branch in a CNN with dilated convolution (DConv-CNN) is independent from the others.

Let us consider the case of Fig~\ref{fig:compare}(b) that dilated convolutions with two different dilation rates (e.g., $r_1=2$ and $r_2=4$) are stacked.
Using the fact that a sequence of $n$ 4-dilated convolutions is ($\mathrm{PGP}^{2}$ + Convolution $\times n$ + $\mathrm{PGP}^{-2}$), this architecture can readily be transformed to the one given in Fig.~\ref{fig:compare}(c) which is the equivalent form with PGP.
We can clearly see that each branch split by the former dilation layer is split again by the latter, and all the branches are eventually rearranged by $\mathrm{PGP}^{-1}$ layers prior the global average pooling (GAP) layer.
In contrast, CNN with PGP (PGP-CNN) is implemented as follows.
The convolution or pooling layer that decreases resolution is set to 1 stride of that layer, after which PGP is inserted (Fig.~\ref{fig:compare}(d)).
In DConv-CNN, all the branches are aggregated through $\mathrm{PGP}^{-1}$ layers just before the GAP layer, which can be regarded as a feature ensemble, while PGP-CNN does not use any $\mathrm{PGP}^{-1}$ layer throughout the network so the intermediate features are not fused until the end of the network.

Unlike feature ensemble (i.e., the case of DConv-CNN), this structure enforces all the branches to learn correct classification results.
As mentioned above, DConv-CNN averages an ensemble of intermediate feature.
The likelihood of each class is then predicted based on the average.
Even if some branches contain useful features (e.g. for image classification) while the other branches do not acquire, the CNN attempts to determine the correct class based on the average, which prevents the useless branches from further learning.
In contrast, thanks to the independency of branches PGP-CNN can perform additional learning with superior regularization.

\subsection{Weight transfer using PGP}
\label{subsection:dil_weight}
\begin{figure}[tb]
	\centering
	\begin{tabular}{ccc}
		\includegraphics[width=.3\textwidth]{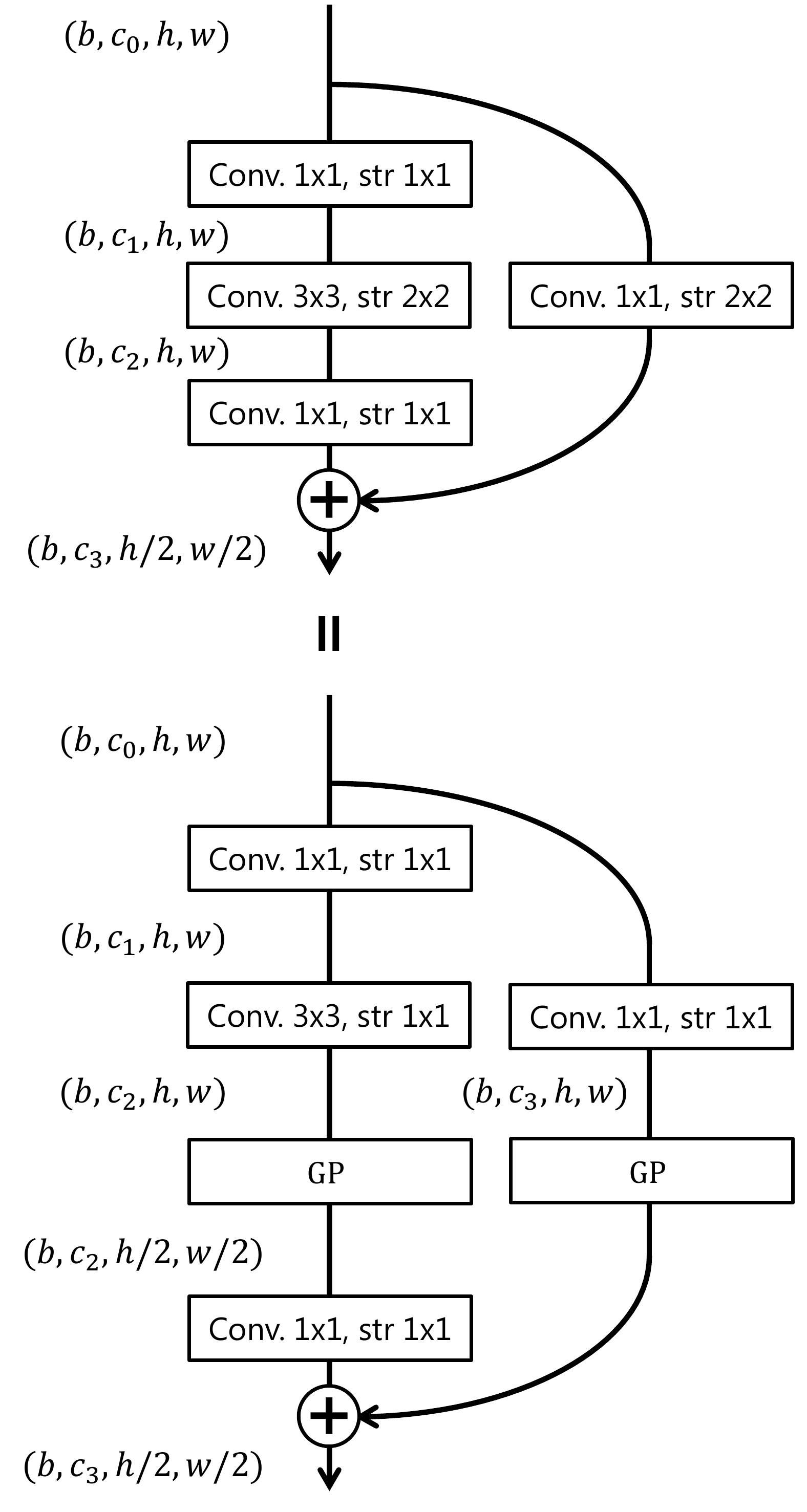} & \includegraphics[width=.3\textwidth]{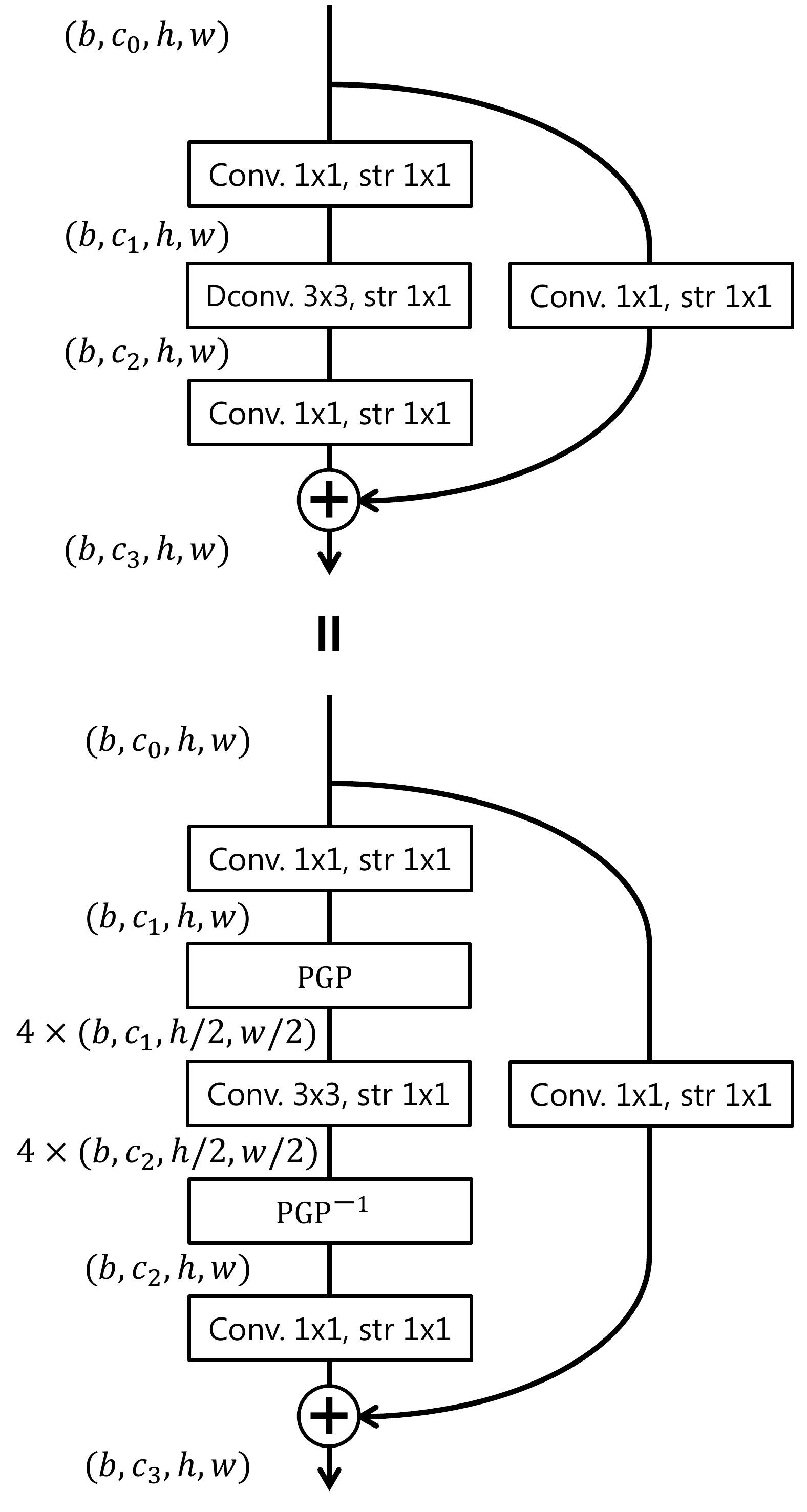} & \includegraphics[width=.3\textwidth]{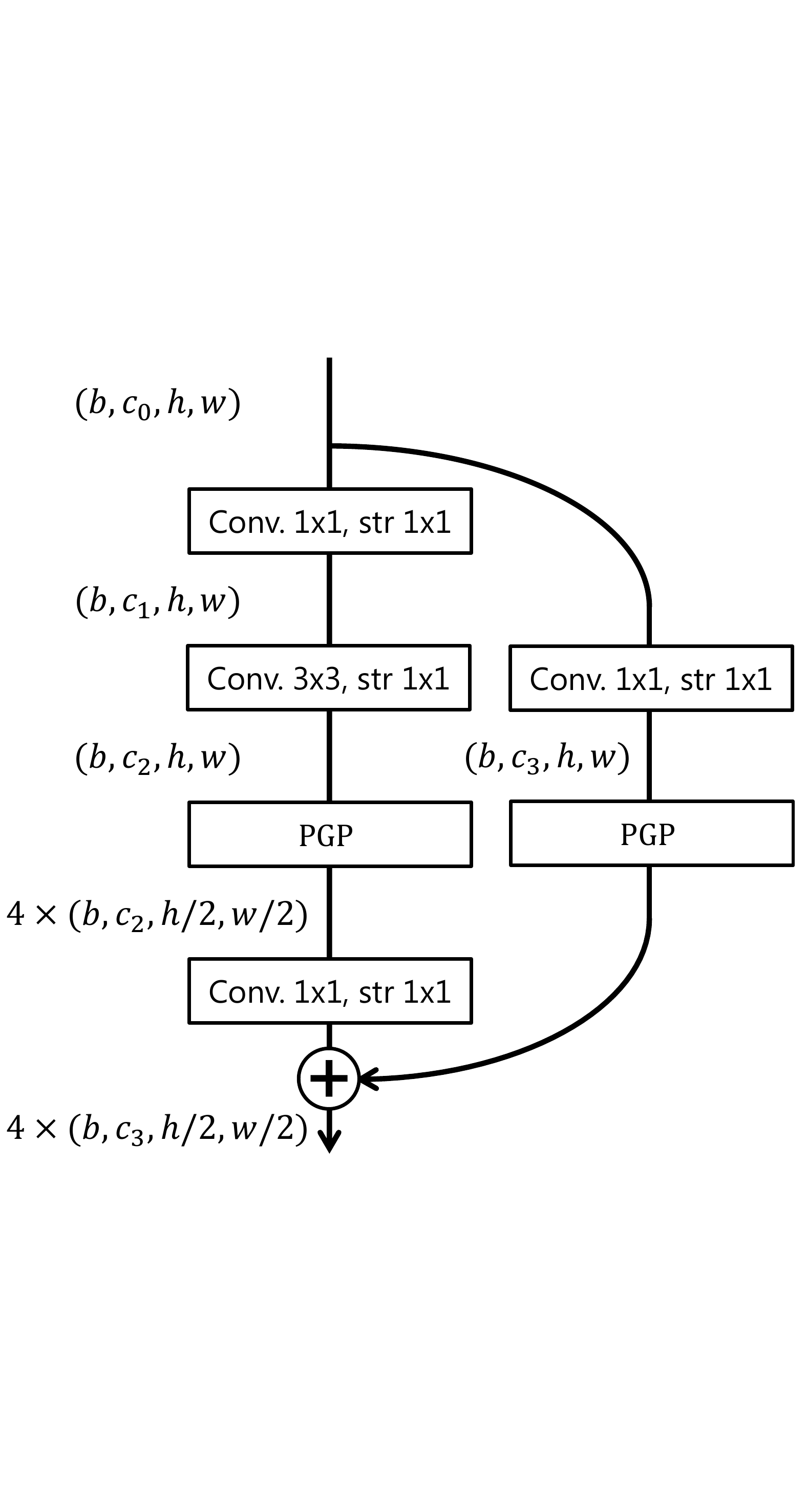} \\
		(a) Base-CNN & (b) DConv-CNN & (c) PGP-CNN \\
	\end{tabular}
	\caption{The structure performing downsampling inside Base-CNN, DConv-CNN, and PGP-CNN. This downsampling structure ($3\times3$ convolution with stride of 2) is used in a derivative of ResNet~\cite{gross2016training}, WideResNet~\cite{zagoruyko2016wide}, and ResNeXt~\cite{xie2017aggregated}.}
	\label{fig:downsample}
\end{figure}

Compared to DConv-CNN, PGP-CNN has a significant advantage in terms of weight transfer.
Specifically, the weights learned on PGP-CNN work well on Base-CNN structure when transferred.
We demonstrate later in the experiments that this can improve performance of the base architecture.
However, DConv-CNN does not have the same structure as Base-CNN because of $\mathrm{PGP}^{-1}$.
The replacement of downsampling by dilated convolutions makes it difficult to transfer weights from the pre-trained base model.
The base CNN with multi-stride convolutions (the top of Fig.~\ref{fig:downsample} (a)) is represented by single-stride convolutions and grid pooling (GP) layers (the bottom of Fig.~\ref{fig:downsample} (a)).
DConv-CNN (the top of Fig.~\ref{fig:downsample} (b)) is equivalent to the base CNN with $\mathrm{PGP}$ and $\mathrm{PGP}^{-1}$ (the bottom of Fig.~\ref{fig:downsample} (b)).
Focusing on the difference in the size of the input features of a $3\times3$ convolution, the size is $(b,c_1,h,w)$ in Base-CNN, whereas $(b,c_1,h/2,w/2)$ in DConv-CNN.
This difference in resolution makes it impossible to directly transfer weights from the pre-trained model.
Unlike DConv-CNN, PGP-CNN maintains the same structure as Base-CNN (Fig.~\ref{fig:downsample} (c)), which results in better transferability and better performance.

This weight transferability of PGP has two advantages.
First, applying PGP to Base-CNN with learned weights in test phase can improve recognition performance without retraining.
For example, if one does not have sufficient computing resources to train PGP-CNN, using PGP only in the test phase can still make full use of the features discarded by downsampling in the training phase to achieve higher accuracy.
Second, the weights learned by PGP-CNN can work well in Base-CNN.
As an example, on an embedded system specialized to a particular CNN, one can not change calculations inside the CNN. 
However, the weights learned by on PGP-CNN can result in superior performance, even when used in Base-CNN at test time.

\section{Experimental Results}
\label{section:experiment}

\subsection{Image classification}
\subsubsection{Datasets}
\label{subsubsection:dataset}
We evaluated PGP on benchmark datasets: CIFAR-10, CIFAR-100~\cite{hinton2007learning}, and SVHN~\cite{netzer2011reading}.
Both CIFAR-10 and CIFAR-100 consist of $32\times32$ color images, with 50,000 images for training and 10,000 images for testing.
The CIFAR-10 dataset has 10 classes, each containing 6,000 images.
There are 5,000 training images and 1,000 testing images for each class.
The CIFAR-100 dataset has 100 classes, each containing 600 images.
There are 500 training images and 100 testing images for each class.
We adopted a standard data augmentation scheme (mirroring/shifting) that is widely used for these two datasets~\cite{he2016deep,huang2016densely,xie2017aggregated}.
For preprocessing, we normalized the data using the channel means and standard deviations.
We used all 50,000 training images for training and calculated the final testing error after training.

The Street View House Numbers (SVHN) dataset consists of $32\times32$ color digit image.
There are 73,257 images in the training set, 26,032 images in the test set, and 531,131 additional images for training.
Following the method in~\cite{zagoruyko2016wide,huang2016densely}, we used all the training data without any data augmentation.

\subsubsection{Implementation details}
\label{subsubsection:detail}
We trained a pre-activation ResNet (PreResNet)~\cite{he2016identity}, All-CNN~\cite{springenberg2014striving} WideResNet~\cite{zagoruyko2016wide}, ResNeXt \cite{xie2017aggregated}, PyramidNet~\cite{han2016deep}, and DenseNet~\cite{huang2016densely} from scratch.
We use a 164-layer network for PreResNet.
We used WRN-28-10, ResNeXt-29 (8x64d), PyramidNet-164 (Bottleneck, alpha=48) and DenseNet-BC-100 (k=12) in the same manner as the original papers.

When comparing the base CNN (\textit{Base}), the CNN with dilated convolutions (\textit{DConv}), and the CNN with PGP (\textit{PGP}), we utilized the same hyperparameter values (e.g. learning rate strategy) to carefully monitor the performance improvements.
We adopted the cosine shape learning rate schedule~\cite{loshchilov2016sgdr}, which smoothly anneals the learning rate~\cite{huang2017snapshot,zoph2017learning}.
The learning rate was initially set to $0.2$ and gradually reduced to $0.002$.
Following \cite{huang2016densely}, for CIFAR and SVHN, we trained the networks using mini-batches of size 64 for 300 epochs and 40 epochs, respectively.
All networks were optimized using stochastic gradient descent (SGD) with a momentum of 0.9 and weight decay of $1.0\times10^{-4}$.
We adopted the weight initialization method introduced in \cite{he2015delving}.
We used the Chainer framework~\cite{chainer_learningsys2015} and ChainerCV~\cite{niitani2017chainercv} for the implementation of all experiments.

\subsubsection{Results}
\label{subsubsection:result_cls}
The results of applying \textit{DConv} and \textit{PGP} on CIFAR-10, CIFAR-100, and SVHN with different networks are listed in Table~\ref{tb:same}.
In all cases, \textit{PGP} outperformed \textit{Base} and \textit{DConv}.

\begin{table}[tb]
	\centering
	\caption{Test errors (\%) with various architectures on CIFAR-10, CIFAR-100, and SVHN. \textbf{DConv:} dilated convolution, \textbf{PGP:} parallel grid pooling.}
	\label{tb:same}
		\begin{tabular}{llcccc} \toprule
			Dataset & Network & \#Params & Base & DConv & PGP \\ \midrule
			\multirow{6}{*}{CIFAR-10} & PreResNet-164 & 1.7M & 4.71$\pm$0.01 & 4.15$\pm$0.12 & \textbf{3.77$\pm$0.09} \\ 
			& All-CNN & 1.4M & 8.42$\pm$0.23 & 8.68$\pm$0.43 & \textbf{7.17$\pm$0.32} \\
			& WideResNet-28-10 & 36.5M & 3.44$\pm$0.06 & 3.88$\pm$0.21 & \textbf{3.13$\pm$0.15} \\
			& ResNeXt-29 (8$\times$64d) & 34.4M & 3.86$\pm$0.16 & 3.87$\pm$0.13 & \textbf{3.22$\pm$0.34} \\
			& PyramidNet-164 ($\alpha$=28) & 1.7M & 3.91$\pm$0.15 & 3.72$\pm$0.07 & \textbf{3.38$\pm$0.09} \\
			& DenseNet-BC-100 (k=12) & 0.8M & 4.60$\pm$0.07 & 4.35$\pm$0.22 & \textbf{4.11$\pm$0.15} \\ \midrule
			\multirow{6}{*}{CIFAR-100} & PreResNet-164 & 1.7M & 22.68$\pm$0.27 & 20.71$\pm$0.27 & \textbf{20.18$\pm$0.45} \\
			& All-CNN & 1.4M & 32.43$\pm$0.31 & 33.10$\pm$0.39 & \textbf{29.79$\pm$0.33} \\
			& WideResNet-28-10 & 36.5M & 18.70$\pm$0.39 & 19.69$\pm$0.50 & \textbf{18.20$\pm$0.17} \\
			& ResNeXt-29 (8$\times$64d) & 34.4M & 19.63$\pm$1.12 & 17.90$\pm$0.57 & \textbf{17.18$\pm$0.29} \\
			& PyramidNet-164 ($\alpha$=28) & 1.7M & 19.65$\pm$0.24 & 19.10$\pm$0.32 & \textbf{18.12$\pm$0.16} \\
			& DenseNet-BC-100 (k=12) & 0.8M & 22.62$\pm$0.29 & 22.20$\pm$0.03 & \textbf{21.69$\pm$0.37} \\ \midrule
			\multirow{6}{*}{SVHN} & PreResNet-164 & 1.7M & 1.96$\pm$1.74 & 1.74$\pm$0.07 & \textbf{1.54$\pm$0.02} \\
			& All-CNN & 1.4M & 1.94$\pm$0.06 & 1.77$\pm$0.07 & \textbf{1.75$\pm$0.02} \\
			& WideResNet-28-10 & 36.5M & 1.81$\pm$0.03 & 1.53$\pm$0.02 & \textbf{1.38$\pm$0.03} \\
			& ResNeXt-29 (8$\times$64d) & 34.4M & 1.81$\pm$0.02 & 1.66$\pm$0.03 & \textbf{1.52$\pm$0.01} \\
			& PyramidNet-164 ($\alpha$=28) & 1.7M & 2.02$\pm$0.05 & 1.86$\pm$0.06 & \textbf{1.79$\pm$0.05} \\
			& DenseNet-BC-100 (k=12) & 0.8M & 1.97$\pm$0.12 & 1.77$\pm$0.06 & \textbf{1.67$\pm$0.07} \\ \bottomrule
		\end{tabular}
\end{table}

\subsubsection{Comparison to other data augmentation methods}
\label{subsubsection:other_aug}
We compare our method to random flipping (RF), random cropping (RC), and random erasing (RE) in Table~\ref{tb:otheraug}.
PGP achieves better performance than any other data augmentation methods when applied alone.
When applied in combination with other data augmentation methods, PGP works in a complementary manner.
Combining RF, RC, RE, and PGP yields an error rate of 19.16\%, which is 11.74\% improvement over the baseline without any augmentation methods.

\begin{table}[tb]
	\centering
	\caption{Test errors (\%) with different data augmentation methods on CIFAR-100 based on PreResNet-164. \textbf{RF:} random flipping, \textbf{RC:} random cropping, \textbf{RE:} random erasing, \textbf{PGP:} parallel grid pooling.}
	\label{tb:otheraug}
	\begin{tabular}{l@{\hspace{3mm}}c@{\hspace{3mm}}c@{\hspace{3mm}}c@{\hspace{3mm}}cc} \toprule
		Network & RF & RC & RE & w/o PGP& w/ PGP \\ \midrule
		\multirow{8}{*}{PreResNet-164} & & & & 30.90$\pm$0.32 & \textbf{22.68$\pm$0.27} \\
		& \checkmark & & & 26.40$\pm$0.18 & \textbf{20.70$\pm$0.34} \\
		& & \checkmark & & 23.94$\pm$0.07 & \textbf{22.34$\pm$0.23} \\
		& & & \checkmark & 27.52$\pm$0.22 & \textbf{21.32$\pm$0.21} \\
		& \checkmark & \checkmark & & 22.68$\pm$0.27 & \textbf{20.18$\pm$0.45} \\
		& \checkmark & & \checkmark & 24.39$\pm$0.55 & \textbf{20.01$\pm$0.14} \\
		& & \checkmark & \checkmark & 22.59$\pm$0.09 & \textbf{20.71$\pm$0.02} \\
		& \checkmark & \checkmark & \checkmark & 21.48$\pm$0.21 & \textbf{19.16$\pm$0.20} \\ \bottomrule
	\end{tabular}
\end{table}

\subsubsection{Comparison to classical ensemble method}
\label{subsubsection:ensemble}
We compare our method to the classical ensemble method in Table~\ref{tb:ensemble}.
On CIFAR-10 and SVHN, a single CNN with the PGP model achieves better performance than an ensemble of three basic CNN models.
Furthermore, PGP works in a complementary manner when applied to the ensemble.

\begin{table}[tb]
	\centering
		\caption{Ensembled test errors (\%) on CIFAR-10/100 and SVHN based on PreResNet-164. ``$\times k$" means an 
			ensembling of $k$ CNN models. \textbf{DConv:} dilated convolution, \textbf{PGP:} parallel grid pooling.}
		\label{tb:ensemble}
		\begin{tabular}{l@{\hspace{3mm}}l@{\hspace{3mm}}cccccc} \toprule
			\multirow{2}{*}{Dataset} & \multirow{2}{*}{Network} & \multicolumn{3}{c}{$\times1$} & \multicolumn{3}{c}{$\times3$} \\ \cmidrule(r){3-8}
			& & Base & DConv & PGP & Base & DConv & PGP \\ \midrule
			CIFAR-10 & \multirow{3}{*}{PreResNet-164} & 4.71 & 4.15 & 3.77 & 3.98 & 3.37 & \textbf{3.14} \\
		 	CIFAR-100 & & 22.68& 20.71 & 20.18 & 19.16 & 17.35 & \textbf{17.15} \\
			SVHN & & 1.96 & 1.74 & 1.54 & 1.65 & 1.51 & \textbf{1.39} \\ \bottomrule
		\end{tabular}
\end{table}

\subsubsection{Weight transfer}
\label{subsubsection:transfer}
The results of applying PGP as a test-time data augmentation method on CIFAR-10, CIFAR-100, and SVHN on different networks are listed in Table~\ref{tb:notrain}.
PGP achieves better performance than the base model, except for on the ALL-CNN and ResNeXt models.
The reason of the incompatibility is left for future work. 

\begin{table}[tb]
		\centering
		\caption{Test errors (\%) when applying PGP as a test-time data augmentation with various architectures on CIFAR-10, CIFAR-100, and SVHN. \textbf{PGP:} parallel grid pooling.}
		\label{tb:notrain}
		\begin{tabular}{llccc} \toprule
			Dataset & Network & \#Params & Base & PGP \\ \midrule
			\multirow{6}{*}{CIFAR-10} & PreResNet-164 & 1.7M & 4.71$\pm$0.01 & \textbf{4.56$\pm$0.14} \\
			& All-CNN & 1.4M & \textbf{8.42$\pm$0.23} & 9.03$\pm$0.30 \\
			& WideResNet-28-10 & 36.5M & 3.44$\pm$0.06 & \textbf{3.39$\pm$0.02} \\
			& ResNeXt-29 (8$\times$64d) & 34.4M & \textbf{3.86$\pm$0.16} & 4.01$\pm$0.21 \\
		 	& PyramidNet-164 ($\alpha$=28) & 1.7M & 3.91$\pm$0.15 & \textbf{3.82$\pm$0.05} \\ 
			& DenseNet-BC-100 (k=12) & 0.8M & 4.60$\pm$0.07 & \textbf{4.53$\pm$0.12} \\ \midrule
			\multirow{6}{*}{CIFAR-100} & PreResNet-164 & 1.7M & 22.68$\pm$0.27 & \textbf{22.19$\pm$0.26} \\
			& All-CNN & 1.4M & \textbf{32.43$\pm$0.31} & 33.24$\pm$0.30 \\
			& WideResNet-28-10 & 36.5M & 18.70$\pm$0.39 & \textbf{18.60$\pm$0.39} \\
			& ResNeXt-29 (8$\times$64d) & 34.4M & \textbf{19.63$\pm$1.12} & 20.02$\pm$0.98 \\
			& PyramidNet-164 ($\alpha$=28) & 1.7M & 19.65$\pm$0.24 & \textbf{19.34$\pm$0.28} \\ 
			& DenseNet-BC-100 (k=12) & 0.8M & 22.62$\pm$0.29 & \textbf{22.33$\pm$0.26} \\ \midrule
			\multirow{6}{*}{SVHN} & PreResNet-164 & 1.7M & 1.96$\pm$0.07 & \textbf{1.83$\pm$0.06} \\
			& All-CNN & 1.4M & \textbf{1.94$\pm$0.06} & 3.86$\pm$0.05 \\
			& WideResNet-28-10 & 36.5M & 1.81$\pm$0.03 & \textbf{1.77$\pm$0.03} \\
			& ResNeXt-29 (8$\times$64d) & 34.4M & \textbf{1.81$\pm$0.02} & 2.47$\pm$0.51 \\
			& PyramidNet-164 ($\alpha$=28) & 1.7M & \textbf{2.02$\pm$0.05} & 2.08$\pm$0.08 \\ 
			& DenseNet-BC-100 (k=12) & 0.8M & 1.97$\pm$0.12 & \textbf{1.89$\pm$0.08} \\ \bottomrule
		\end{tabular}
\end{table}

The results of applying PGP as a training-time data augmentation technique on the CIFAR-10, CIFAR-100, and SVHN datasets with different networks are listed in Table~\ref{tb:tr}.
The models trained with PGP, which have the same structure to the base model at the test phase, outperformed the base model and the model with dilated convolutions.
The models trained with dilated convolutions performed worse compared to the baseline models.

\begin{table}[tb]
		\centering
		\caption{Test errors (\%) when applying PGP as a training-time data augmentation technique with various architectures on CIFAR-10, CIFAR-100, and SVHN. \textbf{DConv:} dilated convolution, \textbf{PGP:} parallel grid pooling.}
		\label{tb:tr}
		\begin{tabular}{llcccc} \toprule
			Dataset & Network & \#Params & Base & DConv & PGP \\ \midrule
			\multirow{6}{*}{CIFAR-10} &PreResNet-164 & 1.7M & 4.71$\pm$0.01 & 7.30$\pm$0.20 & \textbf{4.08$\pm$0.09} \\
			&All-CNN & 1.4M & 8.42$\pm$0.23 & 38.77$\pm$0.85 & \textbf{7.30$\pm$0.31} \\
			&WideResNet-28-10 & 36.5M & 3.44$\pm$0.06 & 7.90$\pm$0.90 & \textbf{3.30$\pm$0.13} \\
			&ResNeXt-29 (8$\times$64d) & 34.4M & 3.86$\pm$0.16 & 16.91$\pm$0.45 & \textbf{3.36$\pm$0.27} \\
			&PyramidNet-164 ($\alpha$=28) & 1.7M & 3.91$\pm$0.15 & 6.82$\pm$0.46 & \textbf{3.55$\pm$0.08} \\
			&DenseNet-BC-100 (k=12) & 0.8M & 4.60$\pm$0.07 & 7.03$\pm$0.70 & \textbf{4.36$\pm$0.10} \\ \midrule
			\multirow{6}{*}{CIFAR-100} &PreResNet-164 & 1.7M & 22.68$\pm$0.27 & 24.90$\pm$0.47 & \textbf{21.01$\pm$0.54} \\
			&All-CNN & 1.4M & 32.43$\pm$0.31 & 64.27$\pm$2.21 & \textbf{30.26$\pm$0.44} \\
			&WideResNet-28-10 & 36.5M & 18.70$\pm$0.39 & 26.90$\pm$0.80 & \textbf{18.56$\pm$0.21} \\
			&ResNeXt-29 (8$\times$64d) & 34.4M & 19.63$\pm$1.12 & 44.57$\pm$13.78 & \textbf{17.67$\pm$0.31} \\
			&PyramidNet-164 ($\alpha$=28) & 1.7M & 19.65$\pm$0.24 & 25.28$\pm$0.30 & \textbf{18.58$\pm$0.13} \\
			&DenseNet-BC-100 (k=12) & 0.8M & 22.62$\pm$0.29 & 27.56$\pm$0.78 & \textbf{22.46$\pm$0.16} \\ \midrule
			\multirow{6}{*}{SVHN} &PreResNet-164 & 1.7M & 1.96$\pm$1.74 & 3.21$\pm$0.32 & \textbf{1.67$\pm$0.04} \\
			&All-CNN & 1.4M & 1.94$\pm$0.06 & 4.64$\pm$0.60 & \textbf{1.80$\pm$0.04} \\
			&WideResNet-28-10 & 36.5M & 1.81$\pm$0.03 & 3.11$\pm$0.17 & \textbf{1.45$\pm$0.03} \\
			&ResNeXt-29 (8$\times$64d) & 34.4M & 1.81$\pm$0.02 & 5.64$\pm$0.61 & \textbf{1.58$\pm$0.03} \\
			&PyramidNet-164 ($\alpha$=28) & 1.7M & 2.02$\pm$0.05 & 3.23$\pm$0.10 & \textbf{1.90$\pm$0.09} \\
			&DenseNet-BC-100 (k=12) & 0.8M & 1.97$\pm$0.12 & 2.74$\pm$0.10 & \textbf{1.78$\pm$0.07} \\ \bottomrule
		\end{tabular}
\end{table}

\subsection{ImageNet classification}
\label{subsubsection:imagenet}
We further evaluated PGP using the ImageNet 2012 classification dataset~\cite{russakovsky2015imagenet}.
The ImageNet dataset is comprised of 1.28 million training images and 50,000 validation images from 1,000 classes.
With the data augmentation scheme for training presented in \cite{gross2016training,xie2017aggregated,chen2017dual}, input images of size $224\times224$ were randomly cropped from a resized image using scale and aspect ratio augmentation.
We reimplemented a derivative of ResNet~\cite{gross2016training} for training.
According to the dilated network strategy~\cite{chen2016deeplab,yu2015multi}, we used dilated convolutions or PGP after the conv4 stage.
All networks were optimized using SGD with a momentum of 0.9 and weight decay of $1.0\times10^{-4}$.
The models were trained over 90 epochs using mini-batches of size 256 on eight GPUs.
We adopted the weight initialization method introduced in \cite{he2015delving}.
We adopted a cosine-shaped learning rate schedule, where the learning rate was initially set to $0.1$ and gradually reduced to $1.0\times10^{-4}$.

The results when applying dilated convolution and PGP on ImageNet with ResNet-50 and ResNet-101 are listed in Table~\ref{tb:imagenet}.
As can be seen in the results, the network with PGP outperformed the baseline performance. 
PGP works in a complementary manner when applied in combination with 10-crop testing and can be used as a training-time or testing-time data augmentation technique.

\begin{table}[tb]
	\centering
	\caption{Classification error (\%) for the ImageNet 2012 validation set. The error was evaluated with and without 10-crop testing for $224\times224$-pixel inputs. \textbf{DConv:} dilated convolution, \textbf{PGP:} parallel grid pooling.}
	\label{tb:imagenet}
	\begin{tabular}{lccccccc} \toprule
		\multirow{2}{*}{Network} & \multirow{2}{*}{\#Params} & \multirow{2}{*}{Train} & \multirow{2}{*}{Test} & \multicolumn{2}{c}{1 crop} & \multicolumn{2}{c}{10 crops} \\ \cmidrule(r){5-8}
		& & & & top-1 & top-5 & top-1 & top-5 \\ \midrule
		\multirow{6}{*}{ResNet-50} & \multirow{6}{*}{25.6M} & Base & Base & 23.69 & 7.00 & 21.87 & 6.07 \\
		& & DConv & DConv & 22.47 & \textbf{6.27} & 21.17 & 5.58 \\
		& & PGP & PGP & \textbf{22.40} & 6.30 & \textbf{20.83} & \textbf{5.56} \\
		& & Base & PGP & 23.32 & 6.85 & 21.62 & 5.93 \\
		& & DConv & Base & 31.44 & 11.40 & 26.79 & 8.19 \\
		& & PGP & Base & 23.01 & 6.66 & 21.22 & 5.74 \\ \midrule
		\multirow{6}{*}{ResNet-101} & \multirow{6}{*}{44.5M} & Base & Base & 22.49 & 6.38 & 20.85 & 5.50 \\
		& & DConv & DConv & \textbf{21.26} & \textbf{5.61} & 20.03 & 5.02 \\
		& & PGP & PGP & 21.34 & 5.65 & \textbf{19.81} & \textbf{5.00} \\
		& & Base & PGP & 22.13 & 6.21 & 20.46 & 5.36 \\
		& & DConv & Base & 25.63 & 8.01 & 22.40 & 6.05 \\
		& & PGP & Base & 21.80 & 5.95 & 20.18 & 5.15 \\ \bottomrule
	\end{tabular}
\end{table}

\subsection{Multi-label classification}
\label{subsubsection:multilabel}
We further evaluated PGP for multi-label classification on the NUS-WIDE~\cite{chua2009nus} and MS-COCO~\cite{lin2014microsoft} datasets.
NUS-WIDE contains 269,648 images with a total of 5,018 tags collected from Flickr.
These images are manually annotated with 81 concepts, including objects and scenes.
Following the method in \cite{zhu2017learning}, we used 161,789 images for training and 107,859 images for testing.
MS-COCO contains 82,783 images for training and 40,504 images for testing, which are labeled as 80 common objects.
The input images were resized to $256\times256$, randomly cropped, and then resized again to $224\times224$.

We trained a derivative of ResNet~\cite{gross2016training}, with pre-training using the ImageNet dataset.
We used the same pre-trained weights throughout all experiments for fair comparison.
All networks were optimized using SGD with a momentum of 0.9 and weight decay of $5.0\times10^{-4}$.
The models were trained for 16 and 30 epochs using mini-batches of 96 on four GPUs for NUS-WIDE and MS-COCO, respectively.
We adopted the weight initialization method introduced in \cite{he2015delving}.
For fine tuning, the learning rate was initially set to $1.0\times10^{-3}$ and divided by 10 after the 8th and 12th epoch for NUS-WIDE and after the 15th and 22nd epoch for MS-COCO.
We employed mean average precision (mAP), macro/micro F-measure, precision, and recall for performance evaluation.
If the confidences for each class were greater than 0.5, the label were predicted as positive.

The results when applying dilated convolution and PGP on ResNet-50 and ResNet-101 are listed in Table~\ref{tb:multilabel}.
The networks with PGP achieved higher mAP compared to the others.
In contrast, the networks with dilated convolutions achieved lower scores than the baseline, which suggests that ImageNet pre-trained weights were not successfully transferred to ResNet with dilated convolutions due to the difference in resolution as we mentioned in Sec.~\ref{subsection:dil_weight}.
This problem can be avoided by maintaining the dilation rate in the layers that perform downsampling, like the method used in a dilated residual networks (DRN)~\cite{yu2017dilated}.
The operation allows the CNN with dilated convolutions to perform convolution with the same resolution as the base CNN. 
The DRN achieved better mAP scores than the base CNN and nearly the same score as the CNN with PGP.
The CNN with PGP was, however, superior to DRN in that the CNN with PGP has the weight transferability.

\begin{table}[tb]
	\centering
	\caption{Quantitative results on NUS-WIDE and MS-COCO. \textbf{DConv:} dilated convolution, \textbf{DRN:} dilated residual networks, \textbf{PGP:} parallel grid pooling.}
	\label{tb:multilabel}
	\begin{tabular}{llccccccccc} \toprule
		\multirow{2}{*}{Dataset} & \multirow{2}{*}{Network} & \multirow{2}{*}{\#Params} & \multirow{2}{*}{Arch.} & \multirow{2}{*}{mAP} & \multicolumn{3}{c}{Macro} & \multicolumn{3}{c}{Micro} \\ \cmidrule(r){6-11}
		& & & & & F & P & R & F & P & R \\ \midrule
	 	\multirow{8}{*}{NUS-WIDE} & \multirow{4}{*}{ResNet-50} & \multirow{4}{*}{25.6M} & Base & 55.7 & 52.1 & 63.3 & \textbf{47.0} & \textbf{70.2} & 75.3 & \textbf{65.8} \\
		& & & DConv & 55.3 & 51.7 & 63.3 & 46.6 & 70.0 & 75.7 & 65.2 \\
		& & & DRN & 55.9 & \textbf{52.2} & 43.8 & \textbf{47.0} & 70.1 & 75.8 & 65.3 \\
		& & & PGP & \textbf{56.1} & 51.9 & \textbf{64.4} & 46.2 & 70.0 & \textbf{76.4} & 64.6 \\ \cmidrule{2-11}
		& \multirow{4}{*}{ResNet-101} & \multirow{4}{*}{44.5M} & Base & 56.5 & 53.0 & 63.6 & 48.3 & 70.4 & 75.4 & 66.1 \\
		& & & DConv & 56.3 & 53.4 & 62.7 & \textbf{49.2} & 70.5 & 74.9 & \textbf{66.5} \\
		& & & DRN & 56.8 & 53.1 & 63.1 & 48.9 & 70.5 & 75.1 & \textbf{66.5} \\
		&& & PGP & \textbf{57.2} & \textbf{53.5} & \textbf{64.6} & 48.7 & \textbf{70.6} & \textbf{75.7} & 66.2 \\ \midrule
	 	\multirow{8}{*}{MS-COCO} & \multirow{4}{*}{ResNet-50} & \multirow{4}{*}{25.6M} & Base & 73.4 & 67.5 & 78.2 & 60.9 & 72.8 & 82.2 & 65.4 \\
	 	& & & DConv & 72.7 & 66.6 & 77.9 & 59.7 & 72.2 & 81.9 & 64.5 \\
	 	& & & DRN & \textbf{74.2} & \textbf{68.3} & 78.8 & \textbf{61.8} & \textbf{73.4} & 82.5 & \textbf{66.1} \\
	 	& & & PGP & \textbf{74.2} & 67.9 & \textbf{80.0} & 60.5 & 73.1 & \textbf{83.7} & 64.9 \\ \cmidrule{2-11}
	 	& \multirow{4}{*}{ResNet-101} & \multirow{4}{*}{44.5M} & Base & 74.6 & 68.9 & 77.6 & 62.8 & 73.8 & 82.2 & 66.9 \\
		& & & DConv & 74.0 & 68.3 & 78.0 & 62.0 & 73.2 & 81.6 & 66.4 \\
		& & & DRN & 75.3 & \textbf{69.7} & 78.2 & \textbf{63.8} & \textbf{74.3} & 81.8 & \textbf{68.1} \\
		&& & PGP & \textbf{75.5} & 69.4 & \textbf{79.8} & 62.9 & 74.2 & \textbf{83.1} & 67.0 \\ \bottomrule
	\end{tabular}
\end{table}

\section{Conclusion}
In this paper, we proposed PGP, which performs downsampling without discarding any intermediate feature and can be regarded as a data augmentation technique.
PGP is easily applicable to various CNN models without altering their learning strategies.
Additionally, we demonstrated that PGP is implicitly used in dilated convolution, which suggests that dilated convolution is also a type of data augmentation technique.
Experiments on CIFAR-10/100, and SVHN with six network models demonstrated that PGP outperformed the base CNN and the CNN with dilated convolutions.
PGP also obtained better results than the base CNN and the CNN with dilated convolutions on the ImageNet dataset for image classification and the NUS-WIDE and MS-COCO datasets for multi-label image classification. 

\clearpage

\bibliographystyle{splncs}
\bibliography{reference}
\end{document}